\pdfoutput=1

\documentclass[11pt]{article}

\usepackage[final]{acl}

\usepackage{times}
\usepackage{latexsym}

\usepackage[T1]{fontenc}

\usepackage[utf8]{inputenc}

\usepackage{microtype}

\usepackage{inconsolata}

\usepackage{graphicx}

\usepackage{makecell}
\usepackage{algorithm}
\usepackage{algpseudocode}
\usepackage{hyperref}
\usepackage{multirow}

\title{Boosting Logical Fallacy Reasoning in LLMs via Logical Structure Tree}

\author{Yuanyuan Lei and Ruihong Huang\\
        Department of Computer Science and Engineering\\
        Texas A\&M University, College Station, TX\\
        \texttt{\{yuanyuan, huangrh\}@tamu.edu}}

\begin{document}
\maketitle
\begin{abstract}


Logical fallacy uses invalid or faulty reasoning in the construction of a statement. Despite the prevalence and harmfulness of logical fallacies, detecting and classifying logical fallacies still remains a challenging task. We observe that logical fallacies often use connective words to indicate an intended logical relation between two arguments, while the argument semantics does not actually support the logical relation. Inspired by this observation, we propose to build a logical structure tree to explicitly represent and track the hierarchical logic flow among relation connectives and their arguments in a statement. Specifically, this logical structure tree is constructed in an unsupervised manner guided by the constituency tree and a taxonomy of connectives for ten common logical relations, with relation connectives as non-terminal nodes and textual arguments as terminal nodes, and the latter are mostly elementary discourse units. We further develop two strategies to incorporate the logical structure tree into LLMs for fallacy reasoning. Firstly, we transform the tree into natural language descriptions and feed the textualized tree into LLMs as a part of the hard text prompt. Secondly, we derive a relation-aware tree embedding and insert the tree embedding into LLMs as a soft prompt. Experiments on benchmark datasets demonstrate that our approach based on logical structure tree significantly improves precision and recall for both fallacy detection and fallacy classification \footnote{The code and data link is: \url{https://github.com/yuanyuanlei-nlp/logical_fallacy_emnlp_2024}}.

\end{abstract}

\section{Introduction}

\begin{figure*}[t]
  \centering
  \includegraphics[width = 6.3in]{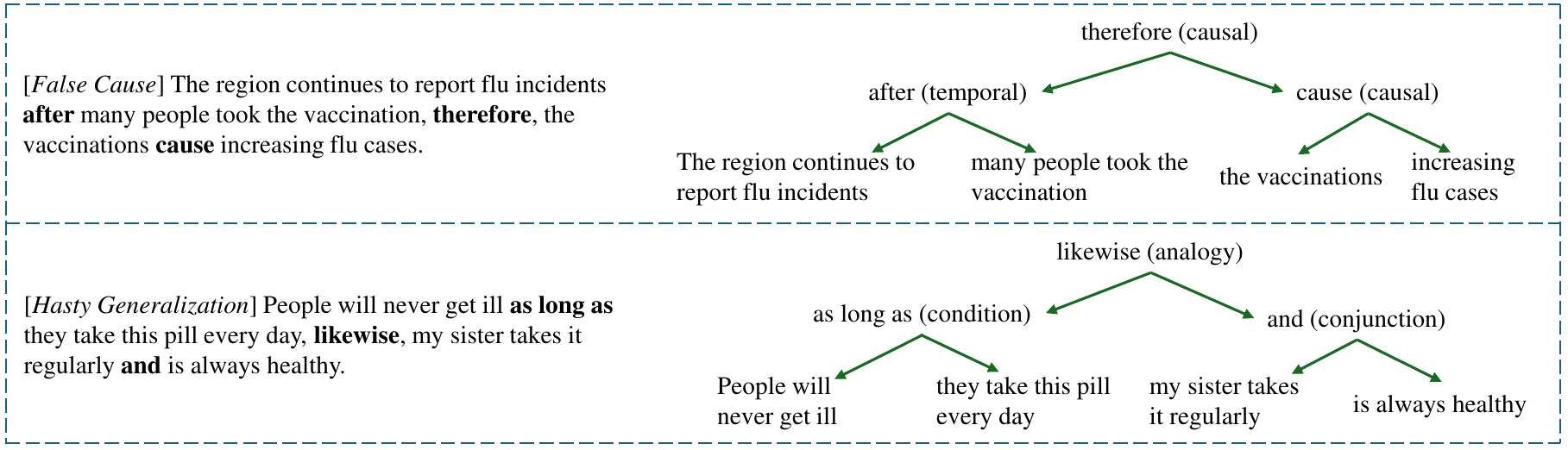}
  \caption{Examples of logical fallacy sentences and their logical structure trees. The logical structure tree features logical relation connectives as non-terminal nodes, and textual arguments as terminal nodes.}
  \label{introduction_example}
\end{figure*}

Logical fallacy refers to the use of invalid or flawed reasoning in an argumentation \cite{risen2007informal, walton2010fallacies, cotton2018argument}. Logical fallacy can occur as unintentional mistakes or deliberate persuasions in a variety of human communications, such as news media \cite{da-san-martino-etal-2019-fine}, educational essay \cite{jin-etal-2022-logical}, political debates \cite{goffredo-etal-2023-argument, mancini-etal-2024-multimodal}, or online discussions \cite{sahai-etal-2021-breaking}. Logical fallacies can lead to harmful consequences for society, such as spreading misinformation \cite{musi2022fallacies, lundy2023tiktok}, raising public health risks \cite{lin2020death}, manipulating public opinions \cite{barclay2018fake, lei-huang-2022-shot, lei-etal-2024-emona}, introducing societal bias and polarization \cite{abd2023relationship}. Despite their prevalence and harmfulness, understanding logical fallacies still remains a challenging task, which requires both semantics understanding and logical reasoning \cite{li-etal-2022-extent, sanyal-etal-2023-apollo}. In this paper, we focus on fallacy detection and classification, and aim to develop an approach that generalizes across different domains and genres.

The key observation is that logical fallacies heavily rely on connective phrases to indicate an intended logical relation between two textual arguments, while the semantics of the arguments do not actually support the claimed logical relation. Figure \ref{introduction_example} shows two examples where the connective phrases were bolded. The first example uses the connective words {\it therefore} and {\it cause} to suggest a causal relation between {\it vaccinations} and {increasing flu cases}, however, the temporal relation between the two events as stated in the first half of the statement does not necessarily entail a causal relation between them, and indeed, their semantics do not actually support the suggested causal relation. Recognizing this discrepancy undermines the credibility of the whole statement. Similarly in the second example, the connective word {\it likewise} is commonly used to indicate an analogy relation, however, the second argument is clearly a specific case of the general condition stated in the first argument and therefore there is no analogy relation between them, and recognizing this mismatch between the suggested logical relation and the real relation enables us to detect this fallacy.

Therefore, we propose to construct a logical structure tree that organizes all connective phrases in a statement and their textual arguments into a hierarchical structure. We expect the logical structure tree to effectively capture the juxtaposition of connective phrase suggested logical relations and the real logical relations between textual arguments, and therefore guide LLMs in fallacy detection and classification. Specifically, a logical structure tree consists of relation connectives as non-terminal nodes and textual arguments as terminal nodes, and the latter mostly corresponds to elementary discourse units (EDU) considered in discourse parsing. Figure \ref{introduction_example} shows the logical structure trees constructed for the two example texts.

As the logical relation indicated by a connective phrase may not be supported by semantics of its arguments in the context, we identify the purposefully indicated logical relations in a context-free unsupervised manner by matching a connective phrase with a taxonomy of connectives compiled for ten common logical relations (conjunction, alternative, restatement, instantiation, contrast, concession, analogy, temporal, condition, causal). To construct a logical structure tree, we first construct a constituency tree for a statement and then search in the constituency tree for connective phrases in the top-down left to right order, and the first found connective phrase will be the root node of the logical structure tree. Next, we identify the text spans of its two arguments using rules and recursively build the left and right sub-trees by applying the same procedure to constituency tree segments corresponding to the two arguments.

The logical structure tree is integrated into LLMs for fallacy reasoning using two strategies. The first considers textualized tree, where we convert the tree into natural language descriptions, making the tree readable by LLMs. Particularly, we describe the relations and arguments in a bottom-up manner, providing the LLMs with insight into logical relations from a local to global perspective. We then concatenate the textualized tree with the instruction prompt, and input them into LLMs as a hard prompt. The second considers tree-based soft prompt, where we derive a relation-aware tree embedding. Specifically, we design relation-specific encoders to process each type of relation and incrementally derive the tree embedding from bottom up to the root node. We then insert the tree embedding into LLMs as a soft prompt for further tuning. Experiments on benchmark datasets across various domains and genres validate that our approach based on logical structure tree effectively improve precision and recall for both fallacy detection and fallacy classification tasks. Our main contributions are summarized as follows:

\begin{itemize}
    \item We propose to construct a logical structure tree to capture the juxtaposition of connective phrase suggested logical relations and the real logical relations between textual arguments, and use it to serve as additional guidance for fallacy detection and classification.
    \item We effectively improve the F1 score for fallacy detection by up to 3.45\% and fallacy classification by up to 6.75\% across various datasets.
\end{itemize}

\section{Related Work}

\noindent\textbf{Logical Fallacy} is erroneous patterns of reasoning \cite{walton1987fallacy, fantino2003logical}. Initial work explored the taxonomy of fallacies \cite{tindale2007fallacies, greenwell2006taxonomy, walton2008argumentation}. Recent works have focused on the automatic detection and classification of fallacies. \citet{habernal-etal-2017-argotario} developed a software that deals with fallacies in question-answering. \citet{sheng-etal-2021-nice} investigated ad hominem fallacy in dialogue responses. \citet{habernal-etal-2018-name} explored the ad hominem fallacy from web argumentations. \citet{stab-gurevych-2017-recognizing} recognized insufficient arguments in argumentation essays. \citet{goffredo2022fallacious} categorized fallacies in political debates. \citet{nakpih2020automated} focused on fallacies in legal argumentations. \citet{musi2022developing} researched fallacies about pandemics on social medias. \cite{alhindi-etal-2022-multitask} proposed a multi-task prompting approach to learn the fallacies from multiple datasets jointly. \citet{jin-etal-2022-logical} proposed a structure-aware method to classify fallacies. Different from \citet{jin-etal-2022-logical} that masked out content words to form a sequence-based pattern, our paper proposes a tree-based hierarchical logical structure to unify both relation connectives and content arguments together.

\vspace{2pt}

\noindent\textbf{Logical Reasoning} abilities of large language models are gaining increasing research attention \cite{xu2023large, chen-etal-2021-neurallog, creswell2022selectioninference, pi2022logigan, jiao-etal-2022-merit, zhou2023leasttomost, sanyal-etal-2023-apollo, parmar-etal-2024-logicbench}. \citet{olausson-etal-2023-linc} combined large language models with first-order logic. \citet{pan-etal-2023-logic, zhang-etal-2023-improved} empowered large language models with symbolic solvers. \citet{pi2022logigan} presented an adversarial pre-training framework to improve logical reasoning. \citet{zhao-etal-2023-explicit} incorporated multi-step explicit planning into the inference procedure. \citet{jiao-etal-2022-merit} proposed a contrastive learning approach to improve logical question-answering. Different from these previous work, we particularly focus on logical fallacy reasoning, aiming to detect and classify fallacies.

\vspace{2pt}

\noindent\textbf{Misinformation} refers to the unverified or false information \cite{guess2020misinformation, armitage2021misinformation, aimeur2023fake, lei-etal-2024-polarity}. Misinformation detection was studied for years, such as fake news \cite{rashkin-etal-2017-truth, lei-huang-2023-identifying, oshikawa-etal-2020-survey}, rumor \cite{ma-etal-2018-rumor, li-etal-2019-rumor}, satire \cite{yang-etal-2017-satirical}, political bias \cite{lei-etal-2022-sentence, feng-etal-2023-pretraining, devatine-etal-2023-integrated, lei-huang-2024-sentence}, propaganda \cite{da-san-martino-etal-2019-fine, da-san-martino-etal-2020-semeval,lei-huang-2023-discourse}. Logical fallacies are often employed within misinformation to present invalid claim as credible, facilitating the spread of misinformation \cite{beisecker2024shades, pauli-etal-2022-modelling, bonial-etal-2022-search}. Developing automatic models to detect logical fallacies can also benefit the identification and mitigation of misinformation.

\begin{table*}[ht]
    \centering
    \scalebox{0.8}{
    \begin{tabular}{|c|c|}
        \hline
        Logical Relations & Relation Connectives \\
        \hline
        conjunction & and, as well as, as well, also, separately \\
        \hline
        alternative & or, either, instead, alternatively, else, nor, neither \\
        \hline
        restatement & \makecell[c]{specifically, particularly, in particular, besides, additionally, in addition, moreover, furthermore,\\plus, not only, indeed, in other words, in fact, in short, in the end, overall, in summary, in details} \\
        \hline
        instantiation & for example, for instance, such as, including, as an example, an as instance, for one thing \\
        \hline
        contrast & \makecell[l]{but, however, yet, while, unlike, rather, rather than, in comparison, by comparison, on the other hand,\\on the contrary, contrary to, in contrast, by contrast, whereas, conversely, not, no, none, nothing, n't} \\
        \hline
        concession & \makecell[c]{although, though, despite, despite of, in spite of, regardless, regardless of, nevertheless,\\nonetheless, even if, even though, even as, even when, even after, even so, no matter} \\
        \hline
        analogy & likewise, similarly, as if, as though, just as, just like, namely \\
        \hline
        temporal & \makecell[l]{during, before, after, when, as soon as, then, next, until, till, meanwhile, in turn, meantime, afterwards,\\simultaneously, at the same time, beforehand, previously, earlier, later, thereafter, finally, ultimately} \\
        \hline
        condition & if, as long as, unless, otherwise, except, whenever, whichever, once, only if, only when, depend on \\
        \hline
        causal & \makecell[c]{because, cause, as a result, result in, due to, therefore, hence, thus, thereby, since, now that,\\consequently, in consequence, in order to, so as to, so that, why, for, accordingly, given, turn out} \\
        \hline
    \end{tabular}}
    \caption{The ten types of logical relations and their relation connectives.}
    \label{logical_relation_connectives}
\end{table*}

\section{Logical Structure Tree}

The logical structure tree consists of relation connectives as non-terminal nodes, and textual arguments as terminal nodes. The relation connectives serve as parent nodes, and the two corresponding arguments are linked as left and right children nodes. Figure \ref{introduction_example} illustrates examples of the logical structure tree. The logical structure tree is constructed in an unsupervised manner, guided by the constituency tree and a taxonomy of connectives complied for ten common logical relations.

\subsection{Relation Connectives}

The logical fallacies usually rely on relation connectives to indicate a logical relation. Inspired by the discourse relations proposed by \citet{prasad-etal-2008-penn}, we define a taxonomy of ten logical relations which are commonly seen: \textit{conjunction}, \textit{alternative}, \textit{restatement}, \textit{instantiation}, \textit{contrast}, \textit{concession}, \textit{analogy}, \textit{temporal}, \textit{condition}, and \textit{causal} relations. Moreover, we build a set of connective words and phrases that correspond to each type of logical relation, as shown in Table \ref{logical_relation_connectives}. This set of connectives includes the explicit discourse connectives from the PDTB discourse relation dataset \cite{prasad-etal-2008-penn}, and is further expanded by manually adding relevant connectives from the development set of the logic fallacy dataset \cite{jin-etal-2022-logical}.

We further conduct a statistical analysis on the distribution of ten logical relations and compare distributions between \textit{fallacy} and \textit{no fallacy} classes as well as across different fallacy classes, with the detailed results shown in Appendix \ref{fallacy_classify_statistics_appendix}. The statistical analysis shows that both the \textit{fallacy} and \textit{no fallacy} classes contain many connective phrases and their distributions of the ten logical relations are also very similar. But as expected, different fallacy types tend to employ varying logical patterns, for example, \textit{False Dilemma} uses more alternative relation, while \textit{Deductive Fallacy} uses more analogy relation.


\subsection{Tree Construction Algorithm}

To construct a logical structure tree $T_{logic}$, we first construct a constituency tree $T_{con}$ for a statement. We use the stanza library\footnote{\url{https://stanfordnlp.github.io/stanza/constituency.html}} to get the constituency tree \cite{qi2020stanza}. At the beginning, $T_{logic}$ is initialized as an empty tree. Then we traverse the constituency tree $T_{con}$ from top to bottom and from left to right, and match relation connectives within each subtree of $T_{con}$. If there is a subtree $S_{con(w)}$ whose text equals to a relation connective $w$, we use the algorithm in section \ref{algorithm_arguments} to extract the two textual arguments $\alpha, \beta$ associated with $w$. Then a new logical subtree $S_{logic(w)}$ is created, with the matched relation connective $w$ as a parent node, and the two arguments $\alpha, \beta$ as its left and right children. This new logical subtree $S_{logic(w)}$ is added into the logical structure tree $T_{logic}$. If the textual arguments $\alpha, \beta$ still contain other relation connectives, then we recursively match relation connectives in the arguments and replace the original argument node in the $T_{logic}$ with the newly created logical subtree. The termination condition is that all the relation connectives in the given text have been matched.

\subsection{Textual Arguments Extraction}
\label{algorithm_arguments}

The textual arguments are the two content components linked by a relation connective. Given a matched relation connective $w$, its corresponding subtree in the $T_{con}$ is $S_{con(w)}$. To extract the arguments of $w$, we find the parent tree of $S_{con(w)}$ in the $T_{con}$, denoted as $P(S_{con(w)})$. The text enclosed by $P(S_{con(w)})$ is the concatenation of all its leaf node texts. If the text enclosed by parent tree $P(S_{con(w)})$ contains content before and after the relation connective $w$, i.e., has the form of $\alpha + w + \beta$, then the left argument of $w$ is $\alpha$ and the right argument is $\beta$. If the text enclosed by parent tree $P(S_{con(w)})$ only contains content after the relation connective $w$, i.e., has the form of $w + \beta$, then the right argument of $w$ is $\beta$, and the left argument $\alpha$ is the text enclosed by grandparent tree $P(P(S_{con(w)}))$ subtracted by the text enclosed by $P(S_{con(w)})$.

\section{Logical Fallacy Reasoning}

\begin{figure*}[t]
  \centering
  \includegraphics[width = 6.3in]{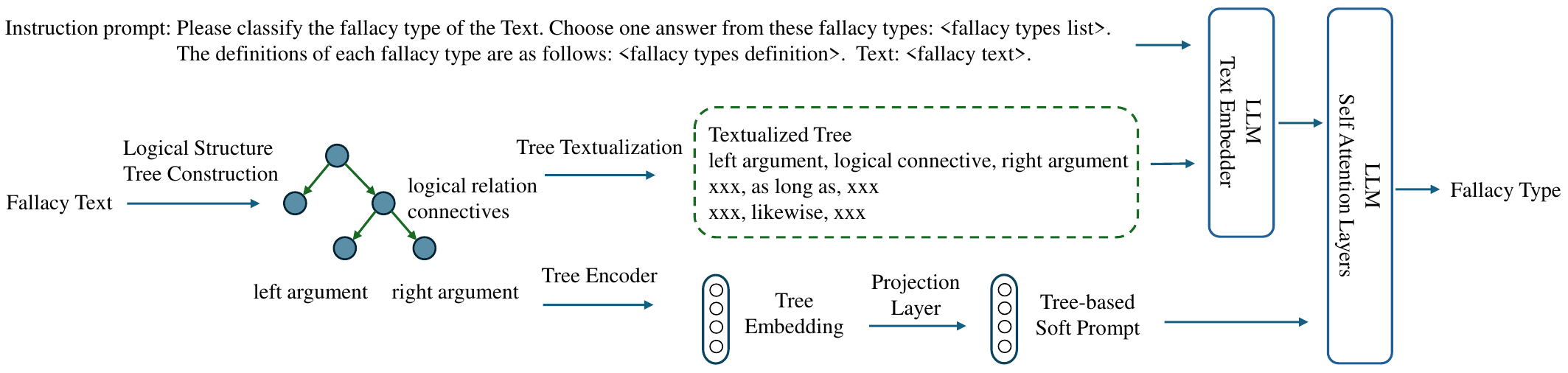}
  \caption{An illustration of logical fallacy classification informed by logical structure tree.}
  \label{methodology_figure}
\end{figure*}

We further design a framework to incorporate the logical structure tree into LLMs for fallacy detection and classification. This framework consists of two main components. The first is textualized tree, where we convert the logical structure tree into natural language descriptions, and feed it into LLMs as a hard text prompt. The second is tree-based soft prompt, where we derive a relation-aware tree embedding, and insert it into LLMs as a soft prompt for additional tuning. The hard and soft prompts are complementary: the hard prompt enriches the instruction with logical structure information, while the soft prompt facilitates direct tuning on tree embeddings. Figure \ref{methodology_figure} shows an illustration.

\subsection{Textualized Tree}

The textualized tree aims to transform the logical structure tree into the textual form, which can be interpretable by LLMs. As shown by the upper path of Figure \ref{methodology_figure}, the textualized tree is represented as a table which consists of three columns: left argument, relation connective, right argument. Each row in the table represents a triplet \textit{(left argument, relation connective, right argument)} corresponding to each logical relation in the tree. In particular, we organize the triplets into the table in a bottom-up order, to provide the LLMs with insight into logical relations from a micro to macro perspective. The textualized tree is then input into the LLMs as a part of the hard text promt:
\begin{equation}
    h_t = TextEmbedder\Bigl(textualize(T_{logic})\Bigr)
\end{equation}
where $textualize(\cdot)$ denotes the textualization operation, $TextEmbedder$ reders to the text embedding layer of LLMs, $h_t$ is the mapped embedding of the textualized tree.

\subsection{Tree-based Soft Prompt}
\label{tree_based_soft_prompt_section}

The tree-based soft prompt is a tree embedding which is projected into LLMs as a soft prompt for further tuning. As shown by the lower path pf Figure \ref{methodology_figure}, this process includes a tree encoder to derive the tree embedding, as well as a projection layer to transform the tree embedding into the same representation space of LLMs.

During the tree encoder stage, we aim to derive a relation-aware tree embedding. To integrate relation information into tree embedding, we design relation-specific encoders to process each type of logical relation. For a simple tree whose children nodes are leaf nodes without hierarchical layers, its embedding is computed as:
\begin{equation}
    e_s = W^r (e_l \oplus e_c \oplus e_r) + b^r
\end{equation}
where $e_s$ is the embedding of this simple tree, $e_l$, $e_c$, $e_r$ are the embeddings of left argument, relation connective, and right argument, which are initialized as the average of word embeddings derived from RoBERTa language model \cite{liu2019roberta}, $\oplus$ denotes feature concatenation, $W^r$, $b^r$ are the trainable parameters of the encoder that corresponds to the relation type $r$, where $W^r \in R^{3d \times d}$, $b^r \in R^d$, and $d=768$ is the dimension of embedding space in RoBERTa. The relation type $r$ is one of the ten logical relations associated with the relation connective.

For the tree with hierarchical structure, we derive the tree embedding incrementally, starting from the bottom simple tree and up towards the root node:
\begin{equation}
    e_t = W^r (\hat{e_l} \oplus e_c \oplus \hat{e_r}) + b^r
\end{equation}
where $e_t$ is the tree embedding, $\hat{e_l}$ is the embedding of the left subtree, $\hat{e_r}$ is the embedding of the right subtree, $e_c$ is the connective embedding.

During the projection stage, we transform the tree embedding $e_t$ into the same representation space of LLMs through a projection layer, which includes two layers of neural networks:
\begin{equation}
    \hat{e_t} = W_2 \Bigl(W_1 e_t + b_1\Bigr) + b_2
\end{equation}
where $W_1$, $W_2$, $b_1$, $b_2$ are the trainable parameters of the projection layer, $W_1\in R^{d \times d'}$, $W_2\in R^{d' \times d'}$, $b_1, b_2 \in R^{d'}$, $d$ is dimension of hidden states in RoBERTa, $d'$ is the dimension of embedding space of the target LLM. $\hat{e_t}$ is the resulting tree-based soft prompt, which is then inserted into LLMs as a token representation within the input sequence.

\subsection{Fallacy Training}

The LLMs take the instruction prompt, textualized tree $h_t$, and tree-based soft prompt $\hat{e_t}$ as input, and generate fallacy label as output. The loss is calculated between the generated text and golden label. The text embedding layer and self attention layers of LLMs are frozen. The tree-based soft prompt $\hat{e_t}$ receives gradients and enables back propagation.

\begin{table*}[t]
    \centering
    \scalebox{0.75}{\begin{tabular}{|l|cccc||cccc||cccc|}
        \hline
        & \multicolumn{4}{c||}{Argotario} & \multicolumn{4}{c||}{Reddit} & \multicolumn{4}{c|}{Climate} \\
        & Precision & Recall & F1 & Acc & Precision & Recall & F1 & Acc & Precision & Recall & F1 & Acc \\
        \hline
        Baselines & & & & & & & & & & & & \\
        \citet{sahai-etal-2021-breaking} & - & - & - & - & 69.57 & 69.27 & 69.20 & - & - & - & - & - \\
        GPT-3.5 & \textbf{92.86} & 14.61 & 25.24 & 41.67 & 54.17 & 15.38 & 23.96 & 50.00 & 70.00 & 7.61 & 13.72 & 33.83 \\
        GPT-3.5 + $T_{logic}$ & 74.72 & 75.55 & 75.14 & 66.16 & 58.26 & 82.94 & 68.45 & 60.61 & \textbf{72.45} & 77.17 & 74.74 & 63.91 \\
        RoBERTa & 81.18 & 83.42 & 82.29 & 75.65 & 65.00 & 76.02 & 70.08 & 66.86 & 67.77 & 89.13 & 76.99 & 63.16 \\
        RoBERTa + $T_{logic}$ & 83.87 & 86.19 & 85.01 & 79.40 & 67.31 & 81.87 & 73.88 & 70.45 & 68.22 & 95.65 & 79.64 & 66.16 \\
        \hline
        Flan-T5 & 81.91 & 85.08 & 83.47 & 77.15 & 67.86 & 77.78 & 72.48 & 69.85 & 68.50 & 94.56 & 79.45 & 66.16 \\
        Flan-T5 + $T_{logic}$ & 84.37 & \textbf{89.50} & 86.86 & 81.65 & 69.31 & 81.87 & 75.07 & 72.24 & 69.17 & \textbf{100.00} & \textbf{81.78} & \textbf{69.17} \\
        Llama-2 & 83.52 & 83.98 & 83.75 & 77.90 & 68.53 & 79.41 & 73.57 & 70.96 & 68.80 & 93.48 & 79.26 & 66.16 \\
        Llama-2 + $T_{logic}$ & 86.02 & 88.40 & \textbf{87.19} & \textbf{82.40} & \textbf{70.05} & \textbf{84.80} & \textbf{76.72} & \textbf{73.73} & 69.17 & \textbf{100.00} & \textbf{81.78} & \textbf{69.17} \\
        \hline
    \end{tabular}}
    \caption{The results of logical fallacy detection on three datasets. The precision, recall, F1 score of \textit{fallacy} class, and accuracy are reported. The rows "+ $T_{logic}$" represent incorporating the logical structure tree into the model.}
    \label{fallacy_detection_results}
\end{table*}

\begin{table}[t]
    \centering
    \scalebox{0.78}{\begin{tabular}{|c|c|c|c|c|c|c|}
        \hline
        Dataset & Train & Dev & Test & Fallacy & Benign & Types \\
        \hline
        Argotario & 863 & 201 & 267 & 909 & 422 & 5 \\
        Reddit & 2313 & 668 & 335 & 1691 & 1625 & 8 \\
        Climate & 436 & 114 & 133 & 477 & 206 & 9 \\
        Logic & 1849 & 300 & 300 & 2449 & - & 13 \\
        \hline
    \end{tabular}}
    \caption{The number of samples in train/dev/test set, the number of fallacy and no fallacy (benign) samples, and the number of fallacy types in each dataset.}
    \label{dataset_statistics}
\end{table}

\section{Experiments}

\subsection{Datasets}

We experiment with four datasets from various domains and genres. Table \ref{dataset_statistics} shows their statistics.

\vspace{2pt}

\noindent\textbf{Argotario} \cite{habernal-etal-2017-argotario} collects fallacies from the general domain question-answering pairs. The dataset includes the following fallacy labels: \textit{Ad Hominem}, \textit{Appeal to Emotion}, \textit{Hasty Generalization}, \textit{Irrelevant Authority}, \textit{Red Herring}, and \textit{No Fallacy}. We use this dataset for both fallacy detection and classification experiments, and follow the dataset splitting method in \citet{alhindi-etal-2022-multitask}.

\vspace{2pt}

\noindent\textbf{Reddit} \cite{sahai-etal-2021-breaking} collects user generated posts from Reddit, and annotates logical fallacies into: \textit{Slippery Slope}, \textit{Irrelevant Authority}, \textit{Hasty Generalization}, \textit{Black-and-White Fallacy}, \textit{Ad Populum}, \textit{Tradition Fallacy}, \textit{Naturalistic Fallacy}, \textit{Worse Problem Fallacy}, and \textit{No Fallacy}. This dataset is used for both fallacy detection and classification.

\vspace{2pt}

\noindent\textbf{Climate} \cite{alhindi-etal-2022-multitask} collects statements from articles in the climate change domain, and annotated the following fallacies: \textit{Evading the Burden of Proof}, \textit{Cherry Picking}, \textit{Red Herring}, \textit{Strawman}, \textit{Irrelevant Authority}, \textit{Hasty Generalization}, \textit{False Cause}, \textit{False Analogy}, \textit{Vagueness}, and \textit{No Fallacy}.

\vspace{2pt}

\noindent\textbf{Logic} \cite{jin-etal-2022-logical} annotates logical fallacies in the educational materials into 13 types including \textit{Ad Hominem}, \textit{Ad Populum}, \textit{False Dilemma}, \textit{False Cause}, \textit{Circular Reasoning}, \textit{Deductive Fallacy}, \textit{Appeal to Emotion}, \textit{Equivocation}, \textit{Fallacy of Extension}, \textit{Faulty Generalization}, \textit{Intentional Fallacy}, \textit{Fallacy of Credibility}, \textit{Fallacy of Relevance}. This dataset does not include \textit{No Fallacy} class and is only used for fallacy classification.

\subsection{Experimental Settings}

To validate our approach, we experiment on two types of language models: a decoder-only model and an encoder-decoder model. For the decoder-only model, we choose the open-source large language model Llama-2 (llama-2-7b-chat-hf) \cite{touvron2023llama}. For the encoder-decoder model, we choose the Flan-T5-large model \cite{chung2022scaling}. Both the models are trained in a generative setting, where they take the instruction and given text as input, and generate a fallacy label as output. The fallacy detection task generates "Yes" or "No" label as output, while the fallacy classification task generates the name of each fallacy type. We follow \citet{alhindi-etal-2022-multitask} to unify the different names of the same fallacy across datasets, such as \textit{False Dilemma} is converted into \textit{Black-and-White Fallacy} since they are the same fallacy. We also follow \citet{alhindi-etal-2022-multitask} to feed the definitions of each fallacy type into the instruction prompt. The details of instruction prompt are explained in Appendix \ref{method_prompt}. The maximum input length is set to be 1024, number of epochs is 10, weight decay is 1e-2, the gradient accumulation step is 4, learning rate for Llama-2 is 3e-4, and learning rate for Flan-T5 is 3e-5. The Llama-2 model is trained with LoRA \cite{hu2021lora}, with rank 8, alpha 16, dropout 0.05, and trainable modules include q\_proj and v\_proj.

\begin{table*}[t]
    \centering
    \scalebox{0.75}{\begin{tabular}{|l|cccc||cccc||cccc|}
        \hline
        & \multicolumn{4}{c||}{Argotario} & \multicolumn{4}{c||}{Reddit} & \multicolumn{4}{c|}{Logic} \\
        & Precision & Recall & F1 & Acc & Precision & Recall & F1 & Acc & Precision & Recall & F1 & Acc \\
        \hline
        Baselines & & & & & & & & & & & & \\
        \citet{jin-etal-2022-logical} & - & - & - & - & - & - & - & - & 55.25 & 63.67 & 58.77 & 47.67 \\
        \citet{sourati2023robust} & - & - & - & - & - & - & - & - & 63.8 & 63.1 & 62.7 & 63.1 \\
        \citet{sourati2023case} & - & - & - & - & - & - & - & - & 66.3 & 66.4 & 65.7 & - \\
        \citet{alhindi-etal-2022-multitask} & - & - & 59 & 59 & - & - & - & - & - & - & 62 & 68 \\
        \citet{sahai-etal-2021-breaking} & - & - & - & - & 62.72 & 55.91 & 58.41 & - & - & - & - & - \\
        GPT-3.5 & 41.65 & 31.32 & 32.48 & 37.02 & 60.35 & 49.22 & 49.81 & 55.62 & 38.14 & 32.58 & 31.30 & 42.28 \\
        GPT-3.5 + $T_{logic}$ & 49.77 & 38.98 & 40.26 & 48.07 & 63.22 & 57.90 & 57.96 & 65.29 & 36.93 & 40.59 & 35.97 & 47.99 \\
        RoBERTa & 57.97 & 55.98 & 55.92 & 57.46 & 71.99 & 70.37 & 70.42 & 70.76 & 62.50 & 59.66 & 60.03 & 64.88 \\
        RoBERTa + $T_{logic}$ & 59.51 & 58.45 & 58.48 & 59.67 & 75.41 & 74.66 & 74.65 & 74.85 & 67.85 & 63.97 & 64.30 & 67.56 \\
        \hline
        Flan-T5 & 60.91 & 57.40 & 58.46 & 58.01 & 76.37 & 76.10 & 76.01 & 76.47 & 65.24 & 63.60 & 63.60 & 69.23 \\
        Flan-T5 + $T_{logic}$ & 65.23 & 62.12 & 62.95 & 62.78 & 81.98 & 81.34 & 81.25 & 81.29 & \textbf{70.90} & 69.14 & 69.37 & 73.49 \\
        Llama-2 & 60.79 & 58.71 & 59.20 & 59.67 & 77.87 & 77.16 & 77.21 & 77.19 & 65.52 & 63.38 & 63.05 & 69.36 \\
        Llama-2 + $T_{logic}$ & \textbf{65.63} & \textbf{63.29} & \textbf{63.92} & \textbf{64.09} & \textbf{84.84} & \textbf{83.68} & \textbf{83.95} & \textbf{83.63} & 70.70 & \textbf{70.03} & \textbf{69.55} & \textbf{74.16} \\
        \hline
    \end{tabular}}
    \caption{The results of logical fallacy classification on three datasets. The macro precision, recall, F1 score, and accuracy are reported. The rows "+ $T_{logic}$" represent incorporating the logical structure tree into the model.}
    \label{fallacy_classification_results}
\end{table*}

\subsection{Baselines}

We compare our models with the baselines listed below. Besides the existing baselines, we also implement several additional baselines based on the GPT and RoBERTa \cite{liu2019roberta} models:

\vspace{3pt}

\noindent\citet{sahai-etal-2021-breaking}: a multi-granularity network is designed that trains sentence-level representation and the token-level representations jointly.

\vspace{3pt}

\noindent\citet{jin-etal-2022-logical}: a structure-aware framework is developed that forms a sequence-based logical pattern for each text by masking out the content words.

\vspace{3pt}

\noindent\citet{sourati2023robust}: a prototype-based reasoning method that injects background knowledge and explainable mechanisms into the language model.

\vspace{3pt}

\noindent\citet{sourati2023case}: a case-based reasoning that retrieves similar cases from external sources based on goals, counterarguments, and explanation etc.

\vspace{3pt}

\noindent\citet{alhindi-etal-2022-multitask}: a multi-task instruction tuning framework that learns the logical fallacies from multiple datasets collaboratively.

\vspace{3pt}

\noindent\textbf{GPT-3.5}: we prompt the gpt-3.5-turbo model to automatically choose one of the fallacy labels for each text, and the prompt is listed in Appendix \ref{gpt_prompt}.

\vspace{3pt}

\noindent\textbf{GPT-3.5 + $T_{logic}$}: guide the gpt-3.5-turbo model to firstly reason the logical structure of each text, and then choose one of the fallacy labels through a chain-of-thought process \cite{wei2023chainofthought}.

\vspace{3pt}

\noindent\textbf{RoBERTa}: the RoBERTa model is used to encode the text and the average of word embedding is used as the text embedding. A classification head is built on top of the text embedding to classify labels.

\vspace{3pt}

\noindent\textbf{RoBERTa + $T_{logic}$}: we concatenate the text embedding with the logical structure tree embedding, and build classification head on top of the combined embedding to predict labels. The tree embedding is derived based on the method in Section \ref{tree_based_soft_prompt_section}.

\subsection{Fallacy Detection}

The fallacy detection task identifies whether a given text contains logical fallacy or not, which is a binary classification task. The precision, recall, and F1 score of the \textit{fallacy} class, as well as the micro F1 score (i.e., accuracy) are used as evaluation metrics. Table \ref{fallacy_detection_results} presents the performance on the Argotario, Reddit, and Climate datasets.

The results demonstrate that incorporating the logical structure tree effectively improves both precision and recall for logical fallacy detection. This observation is consistent for both types of Llama-2 and Flan-T5 models across all the three datasets, which span various domains and genres. Compared to the baselines that lack logical structure information, our approach based on the logical structure tree noticeably enhances the precision and recall, leading to the F1 score increased by up to 3.45\%. This indicates that the logical structure tree is effective in capturing the difference in logical flows between fallacious and benign texts.

Moreover, informing the large language model GPT-3.5-turbo of logical structure information significantly improves fallacy detection under the zero-shot setting, resulting in a substantial improvement in the F1 score. This underscores the importance of integrating the logical structure information into LLMs for fallacy detection. Also, concatenating the logical structure tree embedding with the text embedding in the RoBERTa model also enhances the performance, which proves the usefulness of this logical structure tree embedding. Overall, incorporating the logical structure tree helps improve fallacy detection for various types of models.

\begin{table*}[t]
    \centering
    \scalebox{0.73}{\begin{tabular}{|l|cccc||cccc||cccc|}
        \hline
        & \multicolumn{4}{c||}{Argotario} & \multicolumn{4}{c||}{Reddit} & \multicolumn{4}{c|}{Climate} \\
        Fallacy Detection & Precision & Recall & F1 & Acc & Precision & Recall & F1 & Acc & Precision & Recall & F1 & Acc \\
        \hline
        Llama-2 & 83.52 & 83.98 & 83.75 & 77.90 & 68.53 & 79.41 & 73.57 & 70.96 & 68.80 & 93.48 & 79.26 & 66.16 \\
        + textualized tree & 85.25 & 86.19 & 85.71 & 80.52 & 69.54 & 80.12 & 74.46 & 71.94 & 68.70 & 97.83 & 80.72 & 67.67 \\
        + tree-based soft prompt & 85.11 & \textbf{88.40} & 86.72 & 81.65 & 69.42 & 83.63 & 75.86 & 72.84 & 68.94 & 98.91 & 81.25 & 68.42 \\ 
        + both (full model) & \textbf{86.02} & \textbf{88.40} & \textbf{87.19} & \textbf{82.40} & \textbf{70.05} & \textbf{84.80} & \textbf{76.72} & \textbf{73.73} & \textbf{69.17} & \textbf{100.00} & \textbf{81.78} & \textbf{69.17} \\
        \hline
        \hline
         & \multicolumn{4}{c||}{Argotario} & \multicolumn{4}{c||}{Reddit} & \multicolumn{4}{c|}{Logic} \\
         Fallacy Classification & Precision & Recall & F1 & Acc & Precision & Recall & F1 & Acc & Precision & Recall & F1 & Acc \\
         \hline
        Llama-2 & 60.79 & 58.71 & 59.20 & 59.67 & 77.87 & 77.16 & 77.21 & 77.19 & 65.52 & 63.38 & 63.05 & 69.36 \\
        + textualized tree & 62.63 & 61.32 & 61.86 & 61.67 & 80.98 & 80.71 & 80.45 & 80.59 & 68.71 & 66.09 & 66.38 & 71.24 \\
        + tree-based soft prompt & 64.34 & 61.89 & 62.30 & 62.98 & 82.87 & 82.57 & 82.30 & 82.35 & 68.75 & 68.72 & 67.52 & 72.58 \\
        + both (full model) & \textbf{65.63} & \textbf{63.29} & \textbf{63.92} & \textbf{64.09} & \textbf{84.84} & \textbf{83.68} & \textbf{83.95} & \textbf{83.63} & \textbf{70.70} & \textbf{70.03} & \textbf{69.55} & \textbf{74.16} \\  
        \hline
    \end{tabular}}
    \caption{The results of ablation study. The precision, recall, F1 score of \textit{fallacy} class are reported for fallacy detection (upper rows). The macro precision, recall, F1 score are reported for fallacy classification (lower rows).}
    \label{ablation_study}
\end{table*}

\begin{table*}[t]
    \centering
    \scalebox{0.8}{\begin{tabular}{|l|ccccc|c|}
        \hline
        & Ad Hominem & Emotional & Generalization & Authority & Red Herring & Macro F1 \\
        \hline
        Llama-2 & 60.79 & 67.33 & 55.38 & 63.16 & 49.35 & 59.20 \\
        Llama-2 + $T_{logic}$ & 63.16 & 72.16 & 61.29 & 67.80 & 55.17 & 63.92 \\
        \hline
    \end{tabular}}
    \caption{The F1 score change across each fallacy type of fallacy classification on Argotario dataset. The fallacy types include Ad Hominem, Emotional Language, Hasty Generalization, Irrelevant Authority, and Red Herring.}
    \label{effect_each_type_argotario}
\end{table*}

\begin{table*}[t]
    \centering
    \scalebox{0.7}{\begin{tabular}{|l|cccccccc|c|}
        \hline
        & Slippery & Authority & Generalization & Black-White & Ad Populum & Tradition & Naturalistic & Worse Problem & Macro F1 \\
        \hline
        Llama-2 & 86.96 & 82.05 & 69.57 & 63.41 & 68.29 & 81.82 & 90.00 & 75.56 & 77.21 \\
        Llama-2 + $T_{logic}$ & 88.89 & 92.31 & 77.27 & 65.22 & 82.93 & 87.18 & 95.25 & 82.61 & 83.95 \\
        \hline
    \end{tabular}}
    \caption{The F1 score change across each fallacy type of fallacy classification on Reddit dataset. The fallacy types include Slippery Slope, Irrelevant Authority, Hasty Generalization, Black-and-White Fallacy, Ad Populum, Tradition Fallacy, Naturalistic Fallacy, and Worse Problem Fallacy.}
    \label{effect_each_type_reddit}
\end{table*}

\begin{table*}[t]
    \centering
    \scalebox{0.8}{\begin{tabular}{|l|ccccccc|}
        \hline
        & Ad Hominem & Ad Populum & False Dilemma & False Cause & Circular & Deductive & Emotional \\
        \hline
        Llama-2 & 82.35 & 72.41 & 78.57 & 68.42 & 61.90 & 62.07 & 66.67 \\
        Llama-2 + $T_{logic}$ & 80.46 & 87.50 & 78.57 & 66.67 & 75.68 & 66.67 & 65.22 \\
        \hline
        & Equivocation & Extension & Generalization & Intentional & Authority & Relevance & Macro F1 \\
        \hline
        Llama-2 & 25.00 & 60.00 & 78.13 & 34.48 & 64.71 & 65.00 & 63.05 \\
        Llama-2 + $T_{logic}$ & 44.44 & 72.22 & 81.03 & 38.71 & 68.97 & 78.05 & 69.55 \\
        \hline
    \end{tabular}}
    \caption{The F1 score change across each fallacy type of fallacy classification on Logic dataset. The fallacy types include Ad Hominem, Ad Populum, False Dilemma (Black-and-White Fallacy), False Cause, Circular Reasoning, Deductive Fallacy, Appeal to Emotion (Emotional Language), Equivocation, Fallacy of Extension, Faulty Generalization (Hasty Generalization), Intentional Fallacy, Fallacy of Credibility (Irrelevant Authority), Fallacy of Relevance (Red Herring).}
    \label{effect_each_type_logic}
\end{table*}

\subsection{Fallacy Classification}

The fallacy classification task classifies the fallacy types for the fallacious text, which is a multi-class classification task excluding the \textit{No Fallacy} class. The macro precision, recall, and F1 score, as well as the micro F1 score (i.e., accuracy) are used as evaluation metrics. Table \ref{fallacy_classification_results} shows the results on the Argotario, Reddit, and Logic datasets.

The results demonstrate that integrating the logical structure tree into Llama-2 and Flan-T5 models notably enhances the performance of fallacy classification, with both precision and recall increased. This conclusion is valid across the three datasets from different domains and genres. Compared to the baselines without logical structure tree, our proposed approach significantly improves precision and recall, leading to an increase of up to 6.75\% in the F1 score. This suggests that the logical structure tree effectively distinguishes the different logical patterns used in each fallacy type, and is applicable across various domains and genres.

In addition, our approach based on the logical structure tree outperforms the previous methods that may lack logical relations information. This highlights the necessity to infuse the logical relations into LLMs for fallacy classification. Besides, our approach achieves higher performance than the baselines that overlook content words. This indicates that analyzing content words also plays an essential role in fallacy reasoning. The logical structure tree connects the logical relations and content arguments together to form a cohesive logical structure, representing the hierarchical logical flow and thereby improving fallacy classification.

\subsection{Ablation Study}

The ablation study of the two designed strategies to incorporate the logical structure tree into LLMs is shown in Table \ref{ablation_study}, where we take Llama-2 model as an example. The upper rows show the results of fallacy detection on the three datasets, and the lower rows show the results of fallacy classification.

The results demonstrate that both the textualized tree and tree-based soft prompt brings improvement for fallacy detection and classification across multiple datasets. This proves that the textualized tree and tree-based soft prompt are complementary with each other: the textualized tree enriches the instruction prompt with logical structure information, and the tree-based soft prompt enables direct learning from the tree embedding. Comparing across these two strategies, the soft prompt usually achieves better performance than the hard text prompt, and exhibits higher recall. Combining the two strategies together leads to the best performance, achieving the highest precision and recall.

\subsection{Effect on Different Fallacy Types}

We further analyze the F1 score change across each fallacy type in the fallacy classification task. The Llama-2 model is used as an example to show the performance change before and after incorporating the logical structure tree. Table \ref{effect_each_type_argotario} presents the F1 score change across each fallacy type on Argotario dataset. The performance change across each fallacy type on the Reddit and Logic dataset are shown in the Table \ref{effect_each_type_reddit} and Table \ref{effect_each_type_logic}. We observe that the logical structure tree brings bigger improvements for the fallacy types such as \textit{Red Herring}, \textit{Hasty Generalization}, \textit{Irrelevant Authority}, \textit{Ad Populum}, \textit{Extension Fallacy}, \textit{Equivocation}, \textit{Circular Reasoning} etc. One possible explanation is that these fallacy types usually employ certain logical relations or logical patterns to persuade the readers. However, the performance increase is less noticeable for the fallacy types such as \textit{Appeal to Emotion} and \textit{Ad Hominem}. It may due to the reason that these fallacies rely more on the emotional or sentimental language instead of logical relations.

\section{Limitations}

We have compiled a set of connective words and phrases for the ten logical relations, as detailed in Table \ref{logical_relation_connectives}. While we have included the common connectives in this set, it may not contain all the possible connectives. The logical structure tree that is constructed based on this connective words set demonstrates its usefulness in fallacy reasoning. Future work can be expanding this connectives set and investigating the effects of various connectives.

\section{Conclusion}

This paper detects and classifies fallacies. We propose a logical structure tree to explicitly represent and track the hierarchical logic flow among relation connectives and their arguments. We also design two strategies to incorporate this logical structure tree into LLMs for fallacy reasoning. Extensive experiments demonstrate the effectiveness of our approach based on the logical structure tree.

\section*{Ethical Considerations}

This paper aims to detect and classify logical fallacies. Logical fallacy is the error or flaws in the reasoning, and can occur in various human communications. Logical fallacies can lead to harmful consequences for society, such as spreading misinformation or introducing societal bias. The goal of this research is to understand logical fallacies, so that we can better identify and mitigate them. The release of code, datasets, and model should be used for mitigating logical fallacies, instead of expanding or disseminating the misinformation.

\section*{Acknowledgements}
We would like to thank the anonymous reviewers
for their valuable feedback and input. We gratefully acknowledge support from National Science
Foundation via the award IIS2127746. Portions of this research were conducted
with the advanced computing resources provided
by Texas A\&M High-Performance Research Computing.

\bibliography{custom}

\appendix

\newpage

\section{Statistical Analysis of Logical Relations}
\label{fallacy_classify_statistics_appendix}

Table \ref{fallacy_identify_statistics} presents the ratio of samples that contain the ten logical relations in \textit{fallacy} and \textit{no fallacy} classes, where we take the Argotario \cite{habernal-etal-2017-argotario} and Reddit \cite{sahai-etal-2021-breaking} datasets as examples. Further, Table \ref{fallacy_classify_statistics} shows the ratio of samples that contain the ten logical relations in each fallacy type, where we take the Logic dataset \cite{jin-etal-2022-logical} as an example.


\section{Instruction Prompt for Fallacy Detection and Classification}
\label{method_prompt}

\subsection{Prompt for Fallacy Detection}

The instruction prompt for the Llama-2 or Flan-T5 baseline model is: "The task is to detect whether the Text contains logical fallacy or not. The logical fallacy can be <fallacy name (fallacy definition)>. Please answer Yes if the Text contains logical fallacy, else answer No. Text: <text>. Answer:"

\noindent The instruction prompt that incorporates the textualized tree into the Llama-2 or Flan-T5 model is: "The task is to detect whether the Text contains logical fallacy or not. The logical fallacy can be <fallacy name (fallacy definition)>. The logical relations in the Text are presented in this table: argument 1, logical relation, argument 2 <textualized tree>. Please answer Yes if the Text contains logical fallacy, else answer No. Text: <text>. Answer:"

\subsection{Prompt for Fallacy Classification}

The instruction prompt for the Llama-2 or Flan-T5 baseline model is: "The task is to classify the fallacy type of the Text. Choose one answer from these fallacy types: <fallacy names list>. The definitions of each fallacy type are as follows. <fallacy name: fallacy definition>. Please classify the fallacy type of the Text. Text: <text>. Answer:"

\noindent The instruction prompt that incorporates the textualized tree into the Llama-2 or Flan-T5 model is: "The task is to classify the fallacy type of the Text. Choose one answer from these fallacy types: <fallacy names list>. The definitions of each fallacy type are as follows. <fallacy name: fallacy definition>. The logical relations in the Text are presented in this table: argument 1, logical relation, argument 2 <textualized tree>. Please classify the fallacy type of the Text. Text: <text>. Answer:"

\section{Prompt for GPT-based baselines}
\label{gpt_prompt}

\subsection{Prompt for Fallacy Detection}

The instruction prompt for the gpt-3.5-turbo baseline is: "The task is to detect whether the Text contains logical fallacy or not. The logical fallacy can be <fallacy name (fallacy definition)>. Please answer Yes if the Text contains logical fallacy, else answer No. Text: <text>. Answer:"

\noindent The instruction prompt that incorporates the logical structure into gpt-3.5-turbo model through a chain-of-thought process is: "The task is to detect whether the Text contains logical fallacy or not. The logical fallacy can be <fallacy name (fallacy definition)>. Please answer Yes if the Text contains logical fallacy, else answer No. Let's think step by step. Firstly, explain the logical relations and logical structure in the text. Secondly, choose the answer. Please mimic the output style in the Example. Example: <example text>. Output: Firstly, explain the logical relations and logical structure in the text. <explanation of logical relations in the example>. Secondly, choose the answer. Answer: <fallacy label of the example>.  Text: <text>. Output:"

\subsection{Prompt for Fallacy Classification}

The instruction prompt for the gpt-3.5-turbo baseline is: "The task is to classify the fallacy type of the Text. Choose one answer from these fallacy types: <fallacy names list>. The definitions of each fallacy type are as follows. <fallacy name: fallacy definition>. Please classify the fallacy type of the Text. Text: <text>. Answer:"

\noindent The instruction prompt that incorporates the logical structure into gpt-3.5-turbo model through a chain-of-thought process is: "The task is to classify the fallacy type of the Text. Choose one answer from these fallacy types: <fallacy names list>. The definitions of each fallacy type are as follows. <fallacy name: fallacy definition>. Please classify the fallacy type of the Text. Let's think step by step. Firstly, explain the logical relations and logical structure in the text. Secondly, choose the answer. Please mimic the output style in the Example. Example: <example text>. Output: Firstly, explain the logical relations and logical structure in the text. <explanation of logical relations in the example>. Secondly, choose the answer. Answer: <fallacy label of the example>.  Text: <text>. Output:"

\section{The Names and Definitions of Fallacies}

\subsection{Argotario dataset}

The Argotario dataset \cite{habernal-etal-2017-argotario} includes five fallacy types: Ad Hominem, Appeal to Emotion, Hasty Generalization, Irrelevant Authority, Red Herring. The name of Appeal to Emotion is converted into Emotional Language. The definitions of these fallacy types which are used in the instruction prompt are:
\begin{itemize}
    \item Ad Hominem: the text attack a person instead of arguing against the claims.
    \item Emotional Language: the text arouse non-rational emotions.
    \item Hasty Generalization: the text draw a broad conclusion based on a limited sample of population.
    \item Irrelevant Authority: the text cite an authority but the authority lacks relevant expertise.
    \item Red Herring: the text diverge the attention to irrelevant issues.
\end{itemize}

\subsection{Reddit dataset}

The Reddit dataset \cite{sahai-etal-2021-breaking} includes eight fallacy types and their label names are: Slippery Slope, Irrelevant Authority, Hasty Generalization, Black-and-White Fallacy, Ad Populum, Tradition Fallacy, Naturalistic Fallacy, Worse Problem Fallacy. The definitions of these fallacy types which are used in the instruction prompt are:
\begin{itemize}
    \item Slippery Slope: the text suggest taking a small initial step leads to a chain of related events culminating in significant effect.
    \item Irrelevant Authority: the text cite an authority but the authority lacks relevant expertise.
    \item Hasty Generalization: the text draw a broad conclusion based on a limited sample of population.
    \item Black-and-White Fallacy: the text present two alternative options as the only possibilities.
    \item Ad Populum: the text affirm something is true because the majority thinks so.
    \item Tradition Fallacy: the text argue the action has always been done in the tradition.
    \item Naturalistic Fallacy: the text claim something is good or bad because it is natural or unnatural.
    \item Worse Problem Fallacy: the text justify an issue by arguing more severe issues exists.
\end{itemize}

\subsection{Climate dataset}

The Climate dataset \cite{alhindi-etal-2022-multitask} includes the following fallacy types: Evading Burden of Proof, Cherry Picking, Red Herring, Strawman, False Authority, Hasty Generalization, False Cause, Post Hoc, False Analogy, Vagueness. The name of False Authority is replaced by Irrelevant Authority. The class of Post Hoc is combined into False Cause. The definitions of these fallacy types which are used in the instruction prompt are:
\begin{itemize}
    \item Evading Burden of Proof: the text make a claim without evidence or supporting argument.
    \item Cherry Picking: the text selectively present partial evidence to support a claim.
    \item Red Herring: the text diverge the attention to irrelevant issues.
    \item Strawman: the text distort the claim to another one to make it easier to attack.
    \item Irrelevant Authority: the text cite an authority but the authority lacks relevant expertise.
    \item Hasty Generalization: the text draw a broad conclusion based on a limited sample of population.
    \item False Cause: the text assume two correlated events must also have a causal relation.
    \item False Analogy: the text assume two alike things must be alike in other aspects.
    \item Vagueness: the text use ambiguous words, terms, or phrases.
\end{itemize}

\subsection{Logic dataset}

The Logic dataset \cite{jin-etal-2022-logical} annotates 13 types of fallacy: Ad Hominem, Ad Populum, False Dilemma (Black-and-White Fallacy), False Cause, Circular Reasoning, Fallacy of Logic (Deductive Fallacy), Appeal to Emotion (Emotional Language), Equivocation, Fallacy of Extension (Extension Fallacy), Faulty Generalization (Hasty Generalization), Intentional Fallacy, Fallacy of Credibility (Irrelevant Authority), Fallacy of Relevance (Red Herring). The names in the parenthesis are the replaced names used in the instruction prompt. The definitions of these fallacy types which are used in the instruction prompt are:
\begin{itemize}
    \item Ad Hominem: the text attack a person instead of arguing against the claims.
    \item Ad Populum: the text affirm something is true because the majority thinks so.
    \item Black-and-White Fallacy: the text present two alternative options as the only possibilities.
    \item False Cause: the text assume two correlated events must also have a causal relation.
    \item Circular Reasoning: the end of the text come back to the beginning without having proven itself.
    \item Deductive Fallacy: the text has an error in the logical reasoning.
    \item Emotional Language: the text arouse non-rational emotions.
    \item Equivocation: the text use a key term in multiple senses, leading to ambiguous conclusions.
    \item Extension Fallacy: the text attack an exaggerated version of the opponent’s claim.
    \item Hasty Generalization: the text draw a broad conclusion based on a limited sample of population.
    \item Intentional Fallacy: the text show intentional action to incorrectly support an argument.
    \item Irrelevant Authority: the text cite an authority but the authority lacks relevant expertise.
     \item Red Herring: the text diverge the attention to irrelevant issues.
\end{itemize}



\begin{table*}[ht]
    \centering
    \scalebox{0.75}{
    \begin{tabular}{|c|cccccccccc|}
        \hline
        \% & conjunction & alternative & restatement & instantiation & contrast & concession & analogy & temporal & condition & causal \\
        \hline
        fallacy & 37.96 & 46.72 & 1.46 & 0.73 & 48.91 & 1.46 & 6.57 & 10.95 & 16.06 & 69.34 \\
        no fallacy & 28.13 & 40.63 & 3.13 & 0.00 & 42.19 & 3.13 & 1.56 & 7.81 & 15.63 & 56.25 \\
        \hline
        fallacy & 64.04 & 75.44 & 4.39 & 2.92 & 67.54 & 8.19 & 16.67 & 26.90 & 34.80 & 79.24 \\
        no fallacy & 50.31 & 69.63 & 3.37 & 1.53 & 67.18 & 7.98 & 19.94 & 25.46 & 33.44 & 73.01 \\
        \hline
    \end{tabular}}
    \caption{The ratio (\%) of samples that contain the ten logical relations in \textit{fallacy} and \textit{no fallacy} classes in the development set of Argotario (the first two rows) and Reddit (the latter two rows) datasets.}
    \label{fallacy_identify_statistics}
\end{table*}

\begin{table*}[ht]
    \centering
    \scalebox{0.7}{
    \begin{tabular}{|c|cccccccccc|}
        \hline
        \% & conjunction & alternative & restatement & instantiation & contrast & concession & analogy & temporal & condition & causal \\
        \hline
        Ad Hominem & 30.22 & 60.44 & 0.44 & 0.44 & 64.44 & 2.22 & 7.55 & 12.00 & 12.44 & 76.89 \\
        Ad Populum & 20.88 & 47.46 & 0.63 & 1.89 & 27.21 & 1.89 & 5.06 & 10.76 & 10.12 & 72.15 \\
        False Dilemma & 18.34 & 79.81 & 0.91 & 0.00 & 36.69 & 2.75 & 1.83 & 15.59 & 28.44 & 50.45 \\
        False Cause & 46.74 & 62.72 & 00.00 & 00.00 & 36.68 & 1.18 & 3.55 & 37.87 & 11.24 & 86.98 \\
        Circular Claim & 24.24 & 38.63 & 00.00 & 0.75 & 37.87 & 00.00 & 3.03 & 11.36 & 9.09 & 83.33 \\
        Deductive & 28.09 & 63.63 & 00.00 & 00.00 & 39.67 & 0.82 & 19.83 & 17.35 & 24.79 & 76.03 \\
        Emotional & 41.86 & 59.68 & 2.32 & 0.00 & 50.38 & 1.55 & 7.75 & 18.60 & 28.68 & 63.56 \\
        Equivocation & 42.10 & 71.05 & 00.00 & 00.00 & 63.16 & 7.89 & 5.26 & 31.57 & 28.94 & 76.31 \\
        Extension & 54.71 & 73.58 & 00.00 & 1.88 & 62.26 & 0.94 & 11.32 & 11.32 & 18.86 & 87.73 \\
        Generalization & 38.99 & 52.83 & 0.94 & 0.63 & 39.93 & 1.88 & 8.49 & 23.27 & 31.13 & 69.49 \\
        Intentional & 29.46 & 48.21 & 2.67 & 0.89 & 60.71 & 4.46 & 5.36 & 18.75 & 25.00 & 67.85 \\
        Authority & 39.25 & 66.35 & 2.80 & 4.67 & 41.12 & 2.80 & 3.73 & 7.47 & 16.82 & 84.11 \\
        Relevance & 35.96 & 67.54 & 0.87 & 0.00 & 55.26 & 00.00 & 4.38 & 20.17 & 12.28 & 74.56 \\
        \hline
        Overall & 34.49 & 58.97 & 0.87 & 0.82 & 45.97 & 1.85 & 6.91 & 18.22 & 19.75 & 74.37 \\
        \hline
    \end{tabular}}
    \caption{The ratio (\%) of samples that contain the ten logical relations in each fallacy type in the Logic dataset. The fallacy types include Ad Hominem, Ad Populum, False Dilemma (Black-and-White Fallacy), False Cause, Circular Reasoning, Deductive Fallacy, Appeal to Emotion (Emotional Language), Equivocation, Fallacy of Extension, Faulty Generalization (Hasty Generalization), Intentional Fallacy, Fallacy of Credibility (Irrelevant Authority), Fallacy of Relevance (Red Herring).}
    \label{fallacy_classify_statistics}
\end{table*}

\end{document}